\pdfoutput=1

\documentclass[11pt]{article}

\usepackage[]{naacl2021}

\usepackage{times}
\usepackage{latexsym}
\usepackage{graphicx}
\usepackage{multirow}
\usepackage{float}
\usepackage{booktabs}
\usepackage{ctable}
\usepackage{csquotes}
\usepackage{comment}
\usepackage{enumitem}
\usepackage{amsmath}
\usepackage{caption}
\usepackage{subcaption}

\newcommand{\unsupported}{\texttt{Unsupported}}
\newcommand{\supported}{\texttt{Supported}}
\newcommand{\pplgpt}{$\texttt{PPL}_{\texttt{GPT2-B}}$}
\newcommand{\pplgptM}{$\texttt{PPL}_{\texttt{GPT2-M}}$}
\newcommand{\pplgptL}{$\texttt{PPL}_{\texttt{GPT2-L}}$}
\newcommand{\pplgptXL}{$\texttt{PPL}_{\texttt{GPT2-XL}}$}

\newcommand{\pplbert}{$\texttt{PPL}_{\texttt{BERT-B}}$}
\newcommand{\bertb}{$\texttt{BERT-B}_{\texttt{ft}}$}
\newcommand{\bertl}{$\texttt{BERT-L}_{\texttt{ft}}$}

\newcommand{\roberta}{$\texttt{RoBERTa}_{\texttt{ft}}$}

\newcommand{\covidSocial}{Covid-Social}
\newcommand{\covidScientific}{Covid-Scientific}
\newcommand{\feverMix}{FEVER}

\usepackage{xcolor}

\usepackage[T1]{fontenc}

\usepackage[utf8]{inputenc}

\usepackage{microtype}

\title{Towards Few-Shot Fact-Checking via Perplexity}

\author{
Nayeon Lee$^{\dagger}$\thanks{$^*$ Equal contribution.}\quad Yejin Bang$^{\dagger}$$^*$ \\
\bf Andrea Madotto$^{\dagger}$\quad Madian Khabsa$^\mathsection$\quad Pascale Fung$^{\dagger}$ \\
$^{\dagger}$Hong Kong University of Science and Technology $^\mathsection$Facebook AI \\
\texttt {\{nyleeaa,yjbang\}@connect.ust.hk}
}

\begin{document}
\maketitle
\begin{abstract}
Few-shot learning has drawn researchers' attention to overcome the problem of data scarcity. Recently, large pre-trained language models have shown great performance in few-shot learning for various downstream tasks, such as question answering and machine translation. Nevertheless, little exploration has been made to achieve few-shot learning for the fact-checking task. However, fact-checking is an important problem, especially when the amount of information online is growing exponentially every day. In this paper, we propose a new way of utilizing the powerful transfer learning ability of a language model via a perplexity score. The most notable strength of our methodology lies in its capability in \textit{few-shot} learning. With only two training samples, our methodology can already outperform the Major Class baseline by more than absolute $10\%$ on the F1-Macro metric across multiple datasets. Through experiments, we empirically verify the plausibility of the rather surprising usage of the perplexity score in the context of fact-checking and highlight the strength of our few-shot methodology by comparing it to strong fine-tuning-based baseline models. Moreover, we construct and publicly release two new fact-checking datasets related to COVID-19.
\end{abstract}

\section{Introduction}
Few-shot learning is being actively explored to overcome the heavy dependence on large-scale labeled data that serves as a crucial bottleneck to machine learning models. 
Recently, researchers have explored few-shot learning that leverages the powerful transfer learning ability of pre-trained large language models (LMs) in various NLP tasks. \citeauthor{petroni2019language} demonstrated that an LM serves as a good zero-shot learner on the question-answering task due to its encoded commonsense knowledge. Going further, \citeauthor{brown2020language} illustrated the impressive potential of LMs as strong zero-shot and few-shot learners across translation, commonsense reasoning and natural language inference (NLI). However, little or no exploration has been made on few-shot learning in the fact-checking domain, which is a timely and important task in which data-scarcity is particularly problematic.

Previous works have proposed different ways of leveraging LMs to conduct zero- or few-shot learning. One common approach is to query the LM for the missing token (i.e., ``answer'') for the zero-shot question-answering task \cite{petroni2019language, brown2020language} by transforming questions into a form of statement. Another approach is to adopt an in-context learning approach where the input context of the LM is carefully crafted to control the output. For example, a natural language task instruction (e.g., ``Translate English to French:'') or training sample (e.g., ``sea otter => loutre de mer'') is provided as the context for zero-shot/few shot translation \cite{brown2020language}. 

\begin{figure}[t]
  \centering
  \small
  \includegraphics[width=\linewidth]{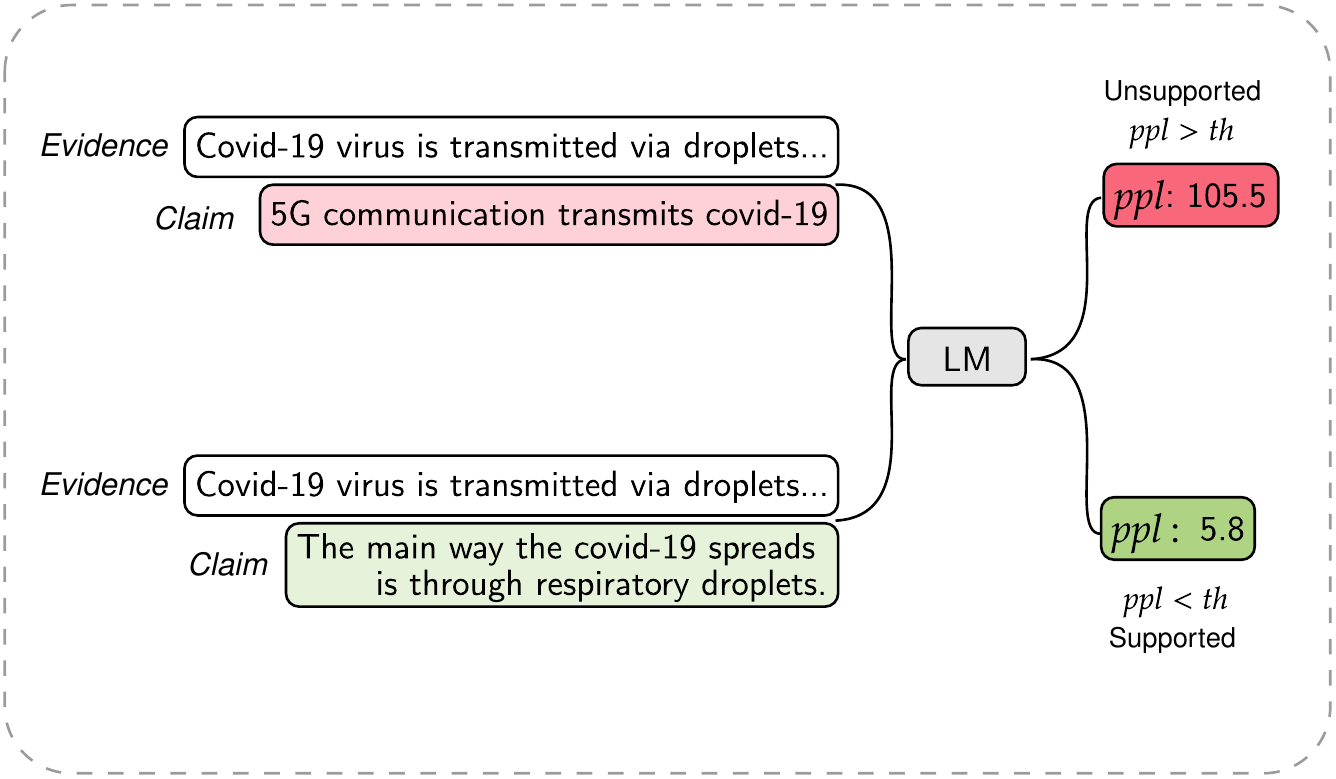}
  \caption{Illustration of our simple yet effective perplexity-based approach. Few-shot data samples are used to find the optimal perplexity threshold $th$ that separates \unsupported~ claims from \supported~ claims. }
  \label{fig:architecture}
\end{figure}

\begin{table*}[t]
\centering
\resizebox{\linewidth}{!}{%
    \begin{tabular}{lclc}
    \toprule
    \textbf{\unsupported~} Claims & Perplexity  & \textbf{\supported~ }Claims & Perplexity\\ \cmidrule(lr){1-2} \cmidrule(lr){3-4}
    5G network can spread diseases. & 826.70  & Beyonce is one of the most famous singers in the world. & 23.03  \\
    All dogs speak English fluently. & 328.23  & Chicago is one of the cities in the United States. & 43.92\\
    Washing hands \textbf{helps} the spread of diseases. &  201.10 & Washing hands \textbf{prevents} the spread of diseases. & 96.74\\
    
    \bottomrule
    \end{tabular}
 } %
\caption{Relations between veracity of claim and perplexity. \unsupported~ claims have higher perplexity compared to \supported~ claims. Note that the perplexity score listed here is using GPT2-base on each of the claims.}
\label{table:ppl_false_true_examples}
\end{table*}

In this work, we explore a new way of leveraging LMs for few-shot learning in the fact-checking task. This is done by leveraging a perplexity score from evidence-conditioned LMs. Fact-checking is the task of verifying a claim based on its corresponding evidence, and one of its most important objectives is to correctly model the relationship between the given claim and evidence. We hypothesize that a perplexity score from evidence-conditioned LMs is helpful for such purpose since perplexity measures the likelihood of a given sentence with reference to previously encountered text (i.e., given the evidence prefix and the LM's training corpus). Therefore, this paper attempts to investigate this hypothesis and proposes a novel perplexity-based few-shot learning methodology for fact-checking.

Through experimental analysis, we empirically demonstrate the effectiveness of our proposed methodology in \textit{few-shot learning}, and we compare it to strong fine-tuning-based baselines. Moreover, we compare different LMs (BERT and GPT2) in different sizes, from small to XL, to unveil interesting insights on which model is more suitable for this task.
Finally, we discuss the potential application of evidence-conditioned perplexity for ranking candidate claims in priority order of the most urgent to be fact-checked to the least.

Our contribution is three-fold: First, we propose an effective way of leveraging the perplexity score in the context of fact-checking. We would like to emphasize that our approach is a simple yet effective way of leveraging large pre-trained LMs. Second, we demonstrate the effectiveness of the perplexity-based approach in the \textit{few-shot} setting by outperforming strong fine-tuned baselines, such as BERT~\cite{devlin2018bert}, RoBERTA~\cite{liu2019roberta}, and XLNet~\cite{yang2019xlnet}, by absolute $10\sim20\%$ F1-Macro scores in the $2$-,$10$-, and $50$-shot settings. Third, we construct two new fact-checking datasets related to COVID-19, which has caused the problem of an ``infodemic''.

\section{Related Work}
\paragraph{Fact-checking} is a complex task that is split into many sub-tasks. First, credible sources of evidence need to be identified. Second, a set of relevant evidence needs to be retrieved from the identified credible sources. Last, veracity classification of claims can be made based on the retrieved evidence. 

Some works have focused on full-pipeline systems that handle all sub-tasks and provide real working web prototypes \cite{karadzhov2017fully,popat2017credibility,popat2018credeye,hasanain2019overview,tokala2019attentivechecker}.
These works use the entire Web as a knowledge source to confirm or reject a claim taking the credibility or reliability of the Web source into account.
Another common setting for fact-checking is to assume a credible evidence source is given (e.g., Wikipedia), and to focus on the evidence retrieval and veracity verification steps only. FEVER~\cite{thorne2018fever} and Tabfact~\cite{chen2019tabfact} are two large datasets for this setting, and there are many follow-up studies working on them ~\cite{yoneda-etal-2018-ucl,nie2019combining,zhong2019reasoning,herzig2020tapas,zhou-etal-2019-gear,hidey-etal-2020-deseption}. 

Our work follows the latter group of works and uses the following setting: given a tuple consisting of claims and relevant evidence, we classify the final fact-checking veracity label of the given claim~\cite{popat2018declare,ma2019sentence,wuevidence}. By doing this, we focus on the methodology for the veracity classification task without worrying about the propagated errors from earlier modules, such as source credibility profiling and evidence retrieval.

\paragraph{Leveraging LMs} as a knowledge base, zero-shot learner or a few-shot learner has been gaining popularity within the NLP field. It was discovered that large pre-trained LMs can store factual knowledge in their parameters \cite{petroni2019language,roberts2020much,madotto2020language}, and that this stored knowledge can help LM to be good at zero-shot and few-shot learning in various NLP tasks, such as question answering, summarization, textual entailment, translation and commonsense reasoning  \cite{brown2020language}.
For the task of fact-checking, \citeauthor{lewis2020retrieval} and \citeauthor{lee2020language} attempted to leverage such LMs. However, they mainly use the model to replace the evidence retriever of the fact-checking pipeline, and they still require training of final veracity classifier. Our work, in contrast, focuses on the few-shot ability of LMs for \textit{veracity classification}.

\section{Preliminary Exploration of Hypothesis}
In this section, we conduct a preliminary investigation to validate the potential of our hypothesis that the perplexity score from an evidence-conditioned LM can provide a signal for claims unsupported by evidence.

For our exploration, we first collect a small set of \supported~ and \unsupported~ claims that can be verified based on the training corpus of the target LM (namely, Wikipedia which is used in the training of many pre-trained LMs). Then, we compare the perplexity scores between them. 

To recap, perplexity is a commonly used metric for measuring the performance of LMs. It is defined as the inverse of the probability of the test set normalized by the number of words:
\begin{equation}
    PPL(X) = \sqrt[n]{\displaystyle\prod_{i=1}^{n} \frac{1}{p(x_{i} |  x_{0}, \ldots, x_{i-1})}}.
    \label{eq:ppl}
\end{equation}
Another way of interpreting perplexity is as a measure of the likelihood of a given test sentence with reference to the training corpus.

From Table~\ref{table:ppl_false_true_examples}, we can observe that  \unsupported~ claims on average have higher perplexity than \supported~ claims. For example, \supported~ claim ``Washing hands prevents the spread of diseases," has a perplexity value of 96.74, whereas the \unsupported~ claim ``All dogs speak English fluently," has a much higher perplexity value of 328.23. We believe these observations support our hypothesis. Thus, we proceed to build our approach based on this hypothesis (Section~\ref{section:methodology}), and conduct experiments (Section~\ref{section:experiments}) and analysis (Section~\ref{section:analysis}) to verify the validity of our perplexity-based fact-checking approach.

\section{Methodology}
\label{section:methodology}
\subsection{Task definition}
In this work, we define our task to be: Given a \{claim, evidence\} pair, determine the veracity of a claim against the evidence - i.e., \supported~ vs. \unsupported~ claims. 
The label \supported~ is assigned when there exists relevant evidence that supports the claim, while \unsupported~ is assigned when there does not exist any supporting evidence. Note that this existence of refuting evidence also places a claim into this latter category.

\subsection{Evidence Conditioned Perplexity}
Although previous works have shown that an LM can encode knowledge from its training corpus, there are a few limitations to solely relying on the pre-trained weights. First, we cannot easily check and guarantee whether the LM has already seen the evidence that is required for verification, and the LM would definitely not have seen the evidence related to newly emerging events after the LM pre-training. For instance, the event of COVID-19 emerged after the release of the GPT2 pre-trained model. Second, although LMs have shown surprising ability in memorizing some knowledge, they are not perfect, as pointed out by previous works~\cite{poerner2019bert,lee2020language}. Therefore, we propose to incorporate evidence into the perplexity calculation by using it as a prefix of the claim. 

There are two popular kinds of LMs: i) unidirectional LMs that are trained with the conventional next token prediction task, and ii) masked LMs that are trained with the masked token prediction token, resulting in a bidirectional LM. We briefly describe how to obtain the evidence-conditioned perplexity for both types of LM:

\paragraph{Unidirectional Language Model Perplexity}

For a unidirectional LM, first we concatenate the evidence and claim to obtain the input to the LM: $X = \{ x_{e_0}, \ldots, x_{e_E}, x_{c_0} \ldots, x_{c_C} \} $, where $E$ and $C$ denote the number of evidence tokens and claim tokens, respectively.
Then, we obtain the evidence-conditioned perplexity by

\begin{align*}
    & PPL(X) \mathalpha{=} \mathalpha{\sqrt[C]{\prod_{i=1}^C \frac{1}{p (x_{c_i} | x_{e_0}, \ldots, x_{e_E}, \ldots, x_{c_{i-1}})}}}.
\end{align*}


Note that the evidence tokens are used to condition the perplexity, yet their conditional probabilities $p(x_{e_i} | x_{e_0}, \ldots, x_{e_{i-1}})$ do not contribute to the $PPL(X)$, which is the main difference from Eq.~(\ref{eq:ppl}).

\paragraph{Masked Language Model Pseudo Perplexity} A masked LM (MLM) is a type of LM, first proposed by \citeauthor{devlin2018bert}, which is trained with the masked token prediction task instead of the next token prediction task. The ``perplexity'' score from the MLM does not mean the same as the conventional perplexity score. Therefore, we use the ``pseudo perplexity'' score proposed by \citeauthor{salazar2019masked}, which is computed by
summing all the log probabilities obtained by sequentially masking each token in the input sentence.

\subsection{Leveraging Perplexity}
Once we obtain the evidence-conditioned perplexity scores for each claim, we find the best threshold $th$ that separates \supported~ claims from \unsupported~ claims. We would like to emphasize that our approach does not involve any parameter update of the LM. We only do inference with the LM, and leverage the few-shot samples as the ``validation set'' to find the optimal single threshold parameter, $th$. Throughout our paper, we refer to our methodology as the ``perplexity-based classifier''.

Given a set of a claim and evidence, if the evidence-conditioned perplexity score is less than the threshold (i.e. $<th$), the claim is \supported~ by the evidence; otherwise it is \unsupported~.

\begin{table}[t]
\small
\centering
\resizebox{\linewidth}{!}{
\begin{tabular}{lccc}
\toprule
Data sets & \begin{tabular}[c]{@{}c@{}}Unsupported \\ claims\end{tabular} & \begin{tabular}[c]{@{}c@{}}Supported \\ claims\end{tabular} & Total \\ \midrule
Covid19-Scientific & 101 & 71 & 172 \\
Covid19-Social & 263 & 77 & 340 \\ 
\feverMix & 3333 & 3333 & 6666 \\ 
\bottomrule
\end{tabular}
}
\caption{Dataset Statistics}
\label{table:data_statistics}
\end{table}

\section{Few-shot Experiment}
\label{section:experiments}

\subsection{Dataset\footnote{Authors from HKUST obtained performed all experiments with the existing datasets and compiled and released the new datasets.}}
All datasets used in the experiment are in English, and we report the data statistics in Table~\ref{table:data_statistics}.

\paragraph{Covid19-Scientific} A new test set is constructed by collecting COVID-19-related myths and scientific truths labelled by reliable sources like MedicalNewsToday, the Centers for Disease Control and Prevention (CDC), and the World Health Organization (WHO). It consists of the most common scientific or medical myths about COVID-19, which must be debunked correctly to ensure the safety of the public (e.g., ``drinking a bleach solution will prevent you from getting COVID-19''). The set contains 172 claims with labels ($\texttt{Supported}$, $\texttt{Unsupported}$) obtained from the aforementioned reliable sources. Note that myths that are unverifiable from current findings are also assigned the $\texttt{Unsupported}$ label.\footnote{Disclaimer: The data were collected during the early outbreak of COVID-19 (March 2020). The veracity may have been updated as the time evolved, but we release the original version of the dataset for future comparison}

The gold evidence is obtained from the winning system of the Kaggle Covid-19 challenge \cite{su2020caire}. This system retrieves the evidence from 59,000 scholarly articles about COVID-19, SARS-CoV-2, and other related corona viruses.\footnote{https://www.kaggle.com/allen-institute-for-ai/CORD-19-research-challenge}

\paragraph{Covid19-Social} Another test set is constructed by crawling 340 COVID-19-related claims fact-checked by journalists from a website called Politifact.com. Unlike the Covid19-Scientific dataset, it contains non-scientific and socially-related claims, such as ``For the coronavirus, the death rate in Texas, per capita of 29 million people, we're one of the lowest in the country.'' Such claims may not be life-and-death matters, but they still have the potential to bring negative sociopolitical effects. Originally, these claims are labelled into six classes \{pants-fire, false, barely-true, half-true, mostly-true, true\}. However, we use it in a binary setup for consistency with the Covid19-Scientific setup by assigning the first three classes to $\texttt{Unsupported}$ and the rest to $\texttt{Supported}$. 

For evidence of each claim, we follow the \citeauthor{alhindi2018your} to obtain the human-written evidence/justification available on the Politifact.com website, from which the claims are crawled. 

\paragraph{FEVER~\cite{thorne2018fever}} Fact Extraction and Verification (FEVER) is a publicly released large-scale dataset generated by altering sentences extracted from Wikipedia to promote research on fact-checking systems. Since our few-shot experiment requires little data, we only leverage the ``Paper Test Dataset'' from the FEVER workshop (https://fever.ai/) resource page to speed up our experiments. 

This dataset originally has three classes, \{Support, Refute, Not Enough Info\}. ``Support" is similar to our \supported~ label, where a claim can be supported by given evidence. ``Refute" is where a claim is "refuted" by given evidence, whereas ``Not Enough Info" means not enough evidence is available for verification.  For our FEVER experiment, we treat ``Refute'' and ``Not Enough Info'' as one class. This is because we believe that in a real scenario both cases are \unsupported~ claims that need attention.

To provide further detail, the ``Support" class is mapped into \supported, and ``Refute''/``Not Enough Info'' is mapped into \unsupported~ to match our task setting. Note that to balance the dataset, we obtain half the data from ``Refute'' and the other half from ``Not Enough Info''. Note that the gold evidence is included in the dataset released by \citeauthor{thorne2018fever}


\begin{table*}[t]
\centering
\resizebox{0.9\linewidth}{!}{%
\begin{tabular}{crcccccccccc}
\toprule
\multicolumn{1}{l}{\multirow{2}{*}{Shot \#}} & \multicolumn{1}{c}{\multirow{2}{*}{Models}} & \multirow{2}{*}{\begin{tabular}[c]{@{}c@{}}Fine-\\ tuning?\end{tabular}}  & \multicolumn{1}{c}{\multirow{2}{*}{Size}} & \multicolumn{2}{c}{Covid-Scientific} & \multicolumn{2}{c}{Covid-Social}  & \multicolumn{2}{c}{\feverMix} \\
\cmidrule(lr){5-6} \cmidrule(lr){7-8} \cmidrule(lr){9-10}
\multicolumn{1}{l}{} & \multicolumn{1}{c}{} &  &  & Acc & F1-Macro & Acc & F1-Macro & Acc & F1-Macro \\ \midrule \midrule

\multicolumn{1}{l}{} & Major Class & N/A & N/A & 58.72\% & 37.00\% & 77.35\% & 43.62\% & \multicolumn{1}{r}{50.00\%} & \multicolumn{1}{r}{33.33\%} \\ \midrule \midrule
\multirow{6}{*}{2} & $\texttt{BERT-B}_{\texttt{ft}}$ & yes & 110M & 47.34\% & 32.21\% & 26.11\% & 23.33\% &  51.56\% &	37.34\% \\
 & $\texttt{BERT-L}_{\texttt{ft}}$ & yes & 336M & 49.39\% & 34.80\% & 37.78\% & 27.81\%& 50.80\%&	36.49\%   \\
  &  $\texttt{RoBERTa}_{\texttt{ft}}$ & yes & 125M & 52.66\% & 34.29\% & 40.75\% & 26.78\% &  50.00\%	& 33.33\% \\
  
 &$\texttt{XLNet}_{\texttt{ft}}$ & yes & 110M & 51.48\% & 48.49\% & 57.67\% & 44.35\% &49.41\%	& 44.65\% \\ \cmidrule(lr){2-10} 
 & \pplgpt & no & 117M & \textbf{66.75\%} & \textbf{64.39\%} & 62.61\% & \textbf{53.61\%} & \textbf{61.92\%}	&\textbf{57.50\%}\\
 & \pplbert & no & 110M & 47.93\% & 38.54\% & \textbf{77.74\%} & 49.15\% & 52.54\% & 41.33\%  \\ \midrule \midrule

\multirow{6}{*}{10} & $\texttt{BERT-B}_{\texttt{ft}}$ & yes & 110M & 46.27\% & 31.70\% & 43.26\% & 30.70\% &   51.56\% &	37.34\%   \\
 & $\texttt{BERT-L}_{\texttt{ft}}$ & yes & 336M & 50.00\% & 36.74\% & 60.49\% & 42.18\% &  50.80\% &	36.49\%  \\
 & $\texttt{RoBERTa}_{\texttt{ft}}$ & yes & 125M & 52.64\% & 40.28\% & 40.73\% & 26.73\% &    50.00\%	& 33.33\%  \\
 & $\texttt{XLNet}_{\texttt{ft}}$ & yes & 110M & 49.69\% & 42.44\% & 59.68\% & 39.45\% &  49.41\%	& 44.65\% \\ \cmidrule(lr){2-10}
 & \pplgpt & no & 117M & \textbf{72.98\%} & \textbf{68.57\%} & \textbf{71.23\%} &\textbf{55.11\%}  &  \textbf{62.82\%} & 57.04\%  \\
 & \pplbert & no & 110M & 63.15\% & 60.77\% & 61.90\% & 46.35\% &  57.59\%	& \textbf{57.11\%}\\ \midrule \midrule
 
\multirow{6}{*}{50} & $\texttt{BERT-B}_{\texttt{ft}}$ & yes & 110M & 56.75\% &	53.61\% & 60.21\% & 36.91\% &  52.18\% &38.82\%  \\
 & $\texttt{BERT-L}_{\texttt{ft}}$ & yes & 336M & 56.75\% & 39.15\% & 64.94\% & 44.07\% & 51.14\%&	39.99\% \\
 & $\texttt{RoBERTa}_{\texttt{ft}}$ & yes & 125M & 56.40\% & 38.97\% & 73.13\% & 45.30\% & 50.44\% & 38.15\%  \\
 & $\texttt{XLNet}_{\texttt{ft}}$ & yes & 110M & 63.22\% & 51.98\% & \textbf{77.62\%} & 43.70\% & 49.18\% &	48.42\%  \\ \cmidrule(lr){2-10} 
 & \pplgpt & no & 117M & \textbf{74.73\%} & \textbf{73.83\%} & 73.63\% & \textbf{59.91\%} & \textbf{67.48\%}	& \textbf{64.70\%} \\
 & \pplbert & no & 110M & 62.53\% & 61.11\% & 71.11\% & 54.72\%  & 57.44\%	& 56.94\%  \\ \bottomrule

\end{tabular}
}
\caption{Results comparison among perplexity-based classifiers and fine-tuned classifiers in 2-shot, 5-shot and 10-shot settings across three different tasks. Models whose names start with $\texttt{PPL}$ are our proposed perplexity-based classifiers. Major Class is a reference to evaluate classifier performance. All test results reported are mean values of three trials with randomly selected n-shot training samples from the dataset, where $n=\{2,10,50\}$.}
\label{table:main_few_shot_results}
\end{table*}

\subsection{Models}
\paragraph{Ours}
We consider one unidirectional LM and one masked LM for our proposed perplexity-based methodology.
\begin{itemize}
    \item \pplgpt -- Our single-parameter classifier based on perplexity from GPT2-base~\cite{radford2019language} (unidirectional LM)
    \item \pplbert -- Our single-parameter classifier based on perplexity from BERT-base~\cite{devlin2018bert} (Masked LM)
\end{itemize}

\paragraph{Baselines} 
We \textit{finetune} various pre-trained Transformer-based~\cite{vaswani2017attention} models to build our baseline classifiers, which is a common approach used to achieve many state-of-the-art results in the literature. 
\begin{itemize}
    \item Major Class -- A simple majority classifier which always assigns the majority class of the training set to all samples. We provide this for reference because some of our dataset classes are imbalanced.
    \item $\texttt{BERT-B}_{\texttt{ft}}$ -- A fine-tuned BERT-base model with a feed-forward classifier trained on top.
    \item $\texttt{BERT-L}_{\texttt{ft}}$ -- A fine-tuned BERT-large model with a feed-forward classifier trained on top.
    \item $\texttt{RoBERTa}_{\texttt{ft}}$ -- A fine-tuned RoBERTa-base model \cite{liu2019roberta} with a feed-forward classifier trained on top. 
    \item $\texttt{XLNet}_{\texttt{ft}}$ -- A fine-tuned XLNet-base model \cite{yang2019xlnet} with a feed-forward classifier trained on top.
\end{itemize}


\subsection{Experimental Setup}
\paragraph{Few-Shot Data Setup} Given $N_{D}$ as the size of the dataset $D$, we do an $n$-shot experiment with $n$ samples from $D$ as a ``validation set'' for our perplexity-based approach or as a ``training set'' for the fine-tuning approach, and the remainder ($N_{D} - n$) as a test set. To give a concrete example, in the 2-shot experiment using the Covid19-Social dataset (340 samples), we have two training samples and 338 test samples. We use three seeds to split the datasets and train the models. For a fair comparison, all the seeds and splits are kept the same across the models.

\paragraph{Evaluation} We mainly evaluate our experiments using accuracy and the Macro-F1 metric. Since some of our datasets are imbalanced (the ratio of \supported~ to \unsupported~ in Table~\ref{table:data_statistics}), we prioritize the overall Macro-F1 score over accuracy.

\paragraph{Training Details} In our methodology, no gradient update is required. Thus, there are no training details such as learning rate, batch size or max-epoch to report. We simply use a small validation set (size of 2,10,50) to find the best-performing hyper-parameter value for the threshold $th$ from the range of $\{0\sim1000\}$. None of the samples from the test set were seen in threshold searching.

For baseline fine-tuned classifiers, we do a grid-search to find the best-performing parameters, as follows: We use a learning rate of $5\mathrm{e}{-6}$ for training the \bertb~, $\texttt{RoBERTa}_{\texttt{ft}}$, and $\texttt{XLNet}_{\texttt{ft}}$ models, while \bertl is trained with a rate of $2\mathrm{e}{-5}$. All models share the same batch size of $32$ and maximum input sequence length of $128$. We also use early-stopping with patience 3 with a maximum of 10 training epochs.
Each experiment is run on an Nvidia GTX 1080 Ti, and each epoch takes $2\sim15$ seconds depending on the number of the training samples $n$. Note that for reproducibility, we will also publicly release the code.

\subsection{Experimental Results}
\label{experimental_results}
Table~\ref{table:main_few_shot_results} reports the few-shot performance of the fine-tuning-based baselines and our perplexity-based classifiers. 

\paragraph{Usage of Perplexity} We can observe that our perplexity-based classifiers, especially \pplgpt~, outperform all Major Class baselines across all tasks in all settings. For instance, \pplgpt~ outperforms the Major Class by a great margin of $16\%$ and $36.8\%$ on accuracy and F1-Macro scores, for the Covid-Scientific dataset in the 50-shot setting. This supports our hypothesis that evidence-conditioned perplexity scores are capable of providing signals regarding the veracity of the given claim. 

Intuitively, we can consider the perplexity score to be mimicking the role of the ``logits'' from a classifier, and we are trying to find the best threshold to map this pseudo-logit-like perplexity score into a veracity label. The classification performance of our perplexity-based approach increases as the shot size increases. As the shot size increases from $2$ to $50$, \pplgpt~ shows an average gain of $8.19\pm2.74\%$ and $7.64\pm1.61\%$ in accuracy and Macro-F1 score, respectively, across all tasks. This is because a greater number of data samples means more anchor perplexity points for threshold searching, and thus, a better threshold to determine the veracity of claims.

\begin{table*}[t]
\centering
\resizebox{0.85\linewidth}{!}{%
\begin{tabular}{lccccccccc}
\toprule
    \multirow{2}{*}{\textbf{LM Type}} & \multirow{2}{*}{\textbf{\begin{tabular}[c]{@{}c@{}}Parameter\\ Size\end{tabular}}} & \multicolumn{2}{c}{\textbf{Covid-Scientific}} & \multicolumn{2}{c}{\textbf{Covid-Social}}  & \multicolumn{2}{c}{\textbf{\feverMix}} \\ 
    \cmidrule(lr){3-4} \cmidrule(lr){5-6} \cmidrule(lr){7-8} 
     &  & \textbf{Acc} & \textbf{F1 Macro} & \textbf{Acc} & \textbf{F1 Macro} & \textbf{Acc} & \textbf{F1 Macro} \\ \midrule
    \pplgpt & 117M & 74.73\% & 73.83\% & 73.63\% & 59.91\% & 67.48\%& 64.70\% \\
    \pplgptM & 345M & 75.11\% & 73.93\% & \textbf{75.43\%} & \textbf{60.23\%}  & 69.02\%	& 66.39\% \\
    \pplgptL & 774M & 76.19\% & 75.53\% & 73.29\% & 59.30\% & 71.66\% & 	69.99\% \\
    \pplgptXL & 1558M & \textbf{78.23\%} & \textbf{77.63\%} & 72.80\% & 59.88\%  & \textbf{73.67\%} & \textbf{71.71\%} \\ 
\bottomrule
\end{tabular}%
}
\caption{Effect of LM parameter size on the performance of proposed perplexity-based approach in 50-shot setting. All the results are the mean value of three trials.}
\label{table:model_size_comparison}
\end{table*}

\begin{table*}[t]
\centering
\resizebox{0.85\linewidth}{!}{%
    \begin{tabular}{llcccccccc}
    \toprule
    
    \multicolumn{1}{c}{\multirow{2}{*}{Shot \#}} & \multirow{2}{*}{Ablation} & \multicolumn{2}{c}{Covid-Scientific} & \multicolumn{2}{c}{Covid-Social}  & \multicolumn{2}{c}{\feverMix} \\ \cmidrule{3-4} \cmidrule(lr){5-6} \cmidrule(lr){7-8}

    \multicolumn{1}{c}{} &  & Acc & F1 Macro & Acc & F1 Macro & Acc & F1 Macro  \\ \midrule
    \multirow{2}{*}{2} 
     & \pplgptXL & \textbf{68.52\%}&\textbf{66.21\%}&\textbf{66.62\%}&\textbf{52.68\%}&\textbf{62.37\%}&\textbf{56.35\%} \\ 
     & $-$ evidence-conditioning & 62.92\% & 59.53\% & 64.32\% & 51.37\% & 54.72\% & 45.65\% \\
     \midrule
    \multirow{2}{*}{50} 
     & \pplgptXL & \textbf{78.23\%} & \textbf{77.63\%} & \textbf{72.80\%} & \textbf{59.88\%} & \textbf{73.67\%} & \textbf{71.71\%}  \\ 
     & $-$ evidence-conditioning & 73.35\% & 70.21\% & 71.97\% & 56.08\% & 56.69\% & 47.58\%\\
    \bottomrule
    \end{tabular} %
}
\caption{Ablation study -- Effect of the evidence-conditioning on the classification performance.}
\label{table:ablation_study}
\end{table*}

\paragraph{Few-shot Comparison to Fine-tuned Baselines} 
Except for the Covid-Social accuracy in the $50$-shot setting, 
both of our proposed classifiers (\pplgpt~, \pplbert~) outperform the fine-tuned baseline classifiers across all tasks in all of the $2$-, $10$- and $50$-shot settings.
For the 2-shot and 10-shot settings, many of the baseline classifiers underperform the Major Class baseline regardless of the task. This implies their failure to learn anything from the fine-tuning step with a limited number of samples. Only after 50-shot do these baselines start to learn and outperform the Major Class baselines. 
This is not surprising, since the pre-trained models are known to perform well in a full-shot scenario, but they do not guarantee good performance when they are shown few samples.

In contrast, our perplexity-based classifiers manage to perform fairly well, even in the 2-shot setting, because our ``classifier'' is a single parameter (i.e., threshold value), which requires no complex learning or optimization. We would like to emphasize that ours consistently outperform the strong Transformer-based baselines across all dataset on the F1-Macro metric by absolute $10\sim20\%$. We argue that these results demonstrate the strength of our approach in low-resource few-shot settings.

\paragraph{BERT vs. GPT2 for Perplexity Scores} Most of the time, \pplgpt~ outperforms \pplbert. For instance, in the 50-shot setting for the \feverMix~ dataset, performance differences are 10.04\% and 7.76\% for accuracy and F1-Macro scores respectively. Based on this observation, we can speculate that the perplexity from a unidirectional LM is more suitable for our proposed method than from a masked LM. This is most likely because the BERT perplexity score is only an estimation based on the ``pseudo-perplexity'' proposed by \citeauthor{salazar2019masked}


\begin{figure*}
     \centering
     \begin{subfigure}[b]{0.3\textwidth}
         \centering
         \includegraphics[width=\textwidth]{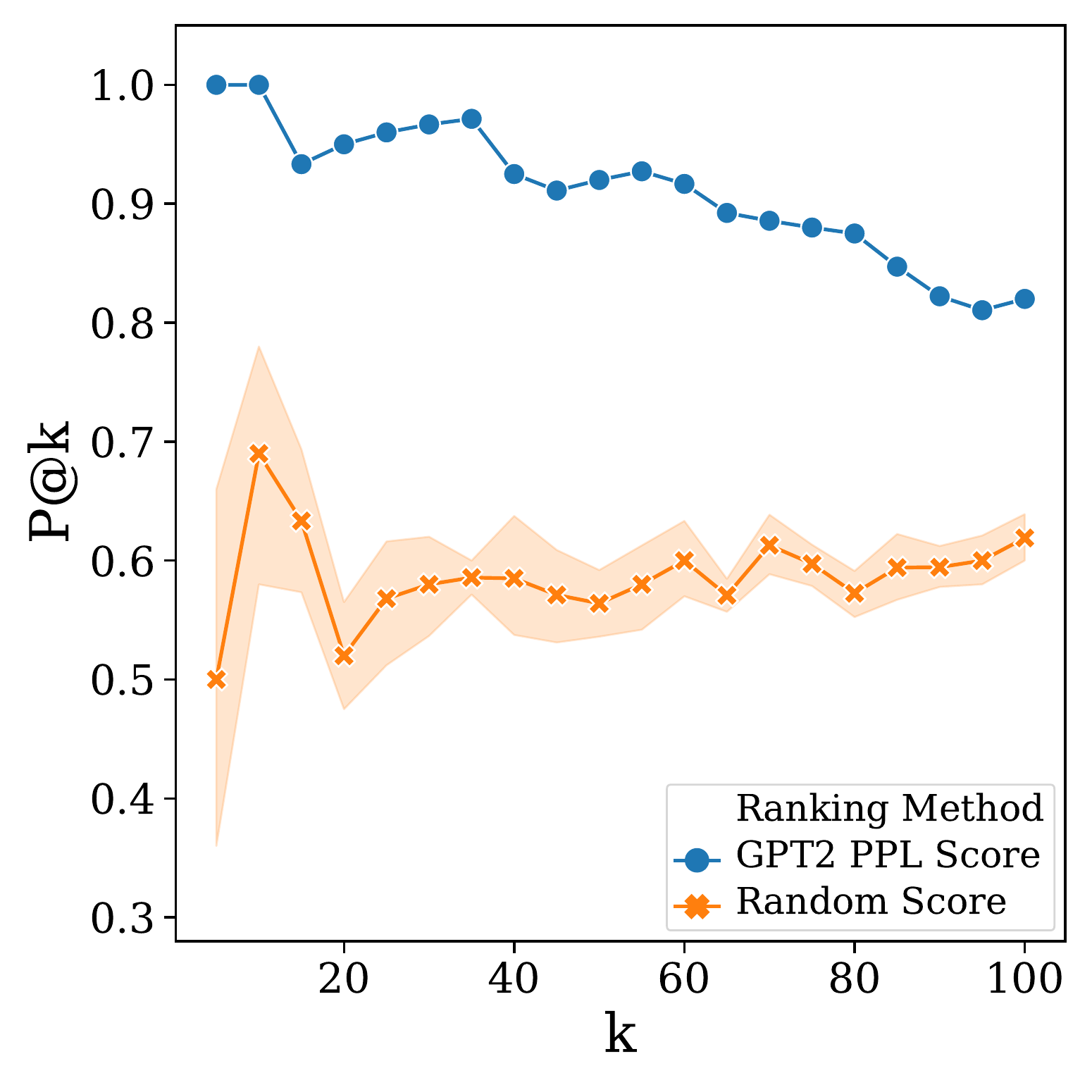}
         \caption{Covid-Scientific}
         \label{fig:covid_scientific_ranking}
     \end{subfigure}
     \hfill
     \begin{subfigure}[b]{0.3\textwidth}
         \centering
         \includegraphics[width=\textwidth]{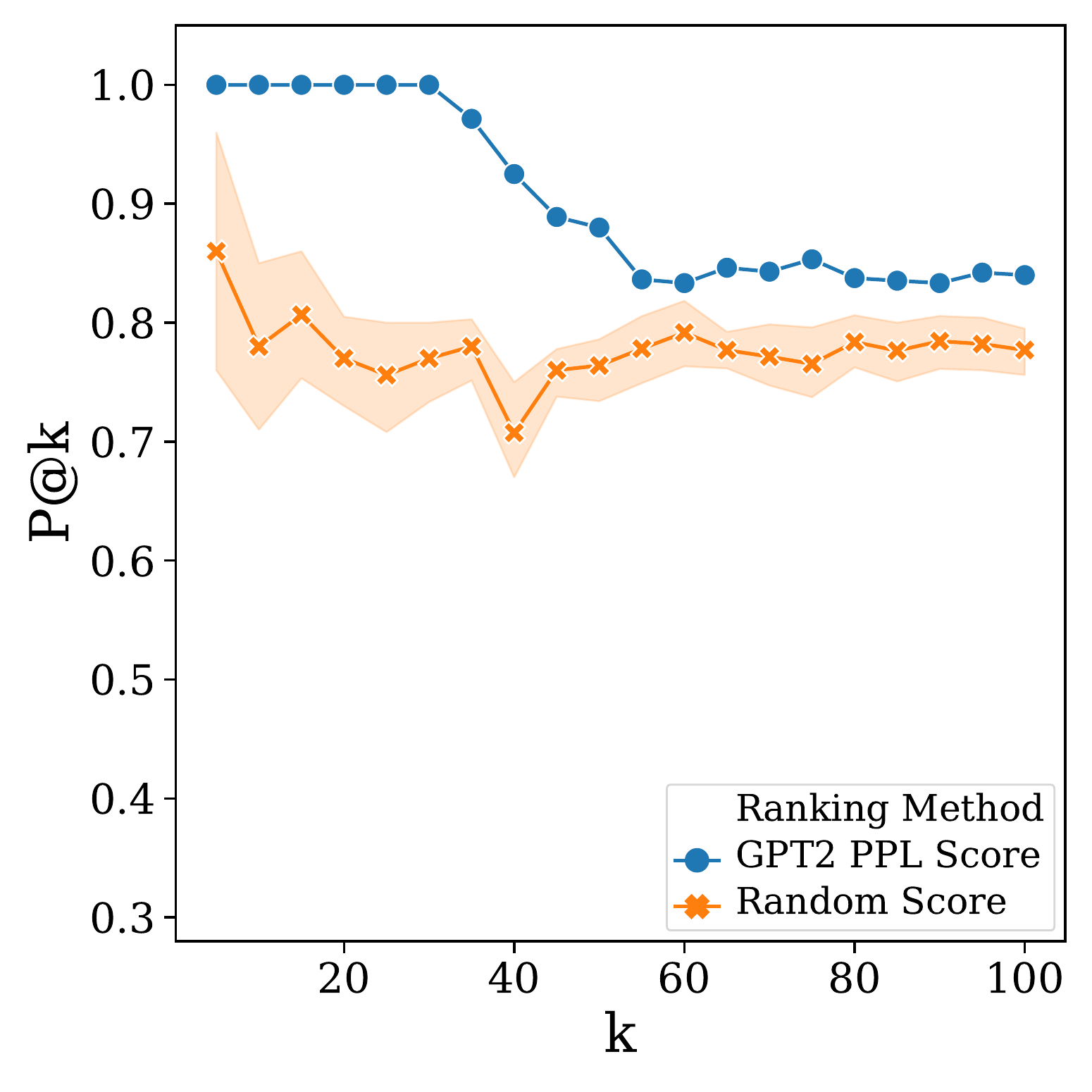}
         \caption{Covid-Social}
         \label{fig:covid_social_ranking}
     \end{subfigure}
     \hfill
     \begin{subfigure}[b]{0.3\textwidth}
         \centering
         \includegraphics[width=\textwidth]{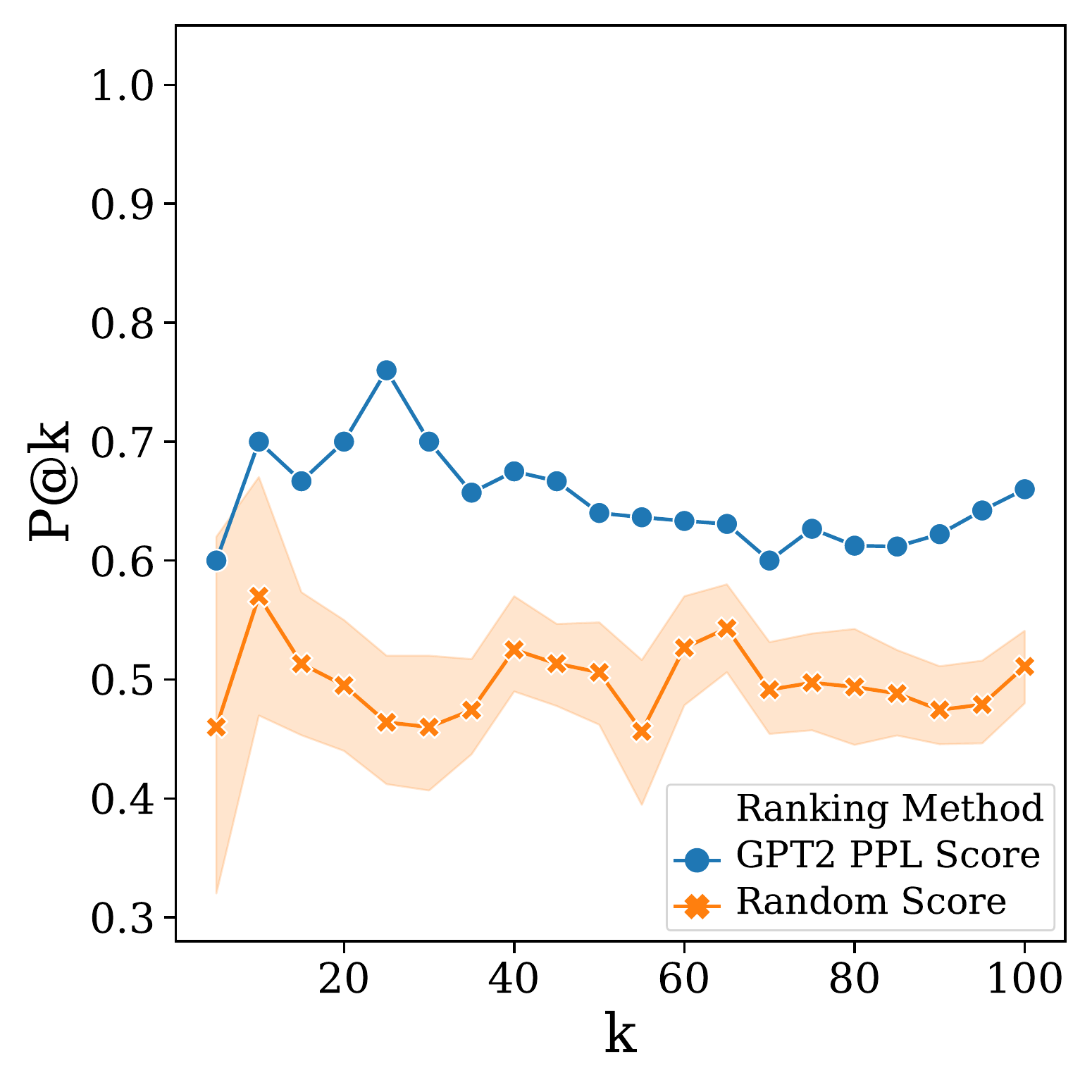}
         \caption{\feverMix}
         \label{fig:fever_ranking}
     \end{subfigure}
        \caption{Precision at top-k. Blue-colored marks indicate precision at each of k when claims are assigned with perplexity scores from GPT2-base to rank claims in reverse order (i.e., higher to lower score). Orange-colored marks indicate mean precision value of 10 trials when random scores are assigned to each claim to rank.
        }
        \label{fig:three graphs}
\end{figure*}

\section{Analysis and Discussion}
\label{section:analysis}
In this section, we conduct multiple analysis to further evaluate and understand aspects of our perplexity-based approach. 


\subsection{Scaling the Language Model Size}
Generally, scaling the model size helps to also improve the model performance, because more parameters mean a stronger learning capability during fine-tuning or training. Also, \citeauthor{roberts2020much} have demonstrated that increasing the parameter size allows for more knowledge to be packed into the LM's parameters. Therefore, we experiment with the model size to see if such findings also extend to our proposed methodology. The following model sizes of GPT2 are investigated: base (\pplgpt), medium (\pplgptM), large (\pplgptL) and xl (\pplgptXL).

Results are reported in Table~\ref{table:model_size_comparison}. As expected, we can observe the trend that the performance increases with parameter size. For instance, \pplgptXL~ is the best performing compared to the other, smaller, models for \covidScientific~ and \feverMix, achieving the new state-of-the-art few-shot results by gaining absolute $\sim4\%$ on \covidScientific~ and  $\sim2\%$ on \feverMix for accuracy/F1-Macro.

\subsection{Ablation Study} 
We carry out an ablation study on the effect of evidence-conditioning in respect of the final perplexity scores and the corresponding final classification performance.
In Table~\ref{table:ablation_study}, we can observe the performance drops when evidence-conditioning is ablated -- the biggest drop is $\sim15\%$ on F1-Macro for the \feverMix~ task in the 50-shot setting. This implies that the perplexity score is assigned \textit{in relation} to the context of the provided evidence.

\subsection{Negation Analysis}
In fact-checking, negation is one of the most difficult challenges, and many state-of-the-art models are brittle against it. \citeauthor{thorne2019adversarial} show that the winning fact-checking systems from the FEVER workshop are brittle against negations, experiencing a huge performance drop when given negated test sets, up to absolute $-29\%$ in accuracy. Therefore, we also conduct analysis regarding the negation handling of our proposed methods by augmenting our dataset with negated examples. 

\paragraph{Template-based Data Negation} We create our negated dataset by replacing all the auxiliary verbs (e.g., is, can) with their corresponding negated forms (e.g., is not, can not), and vice versa. We apply this approach to the \covidScientific~ dataset and obtain a new version that contains \{original-sentence ($S_{\mathrm{original}}$), negated-sentence ($S_{\mathrm{negated}}$)\} pairs. Note that the evidence is kept the same, but the veracity label of $S_{\mathrm{original}}$ is negated (i.e., \supported~ is negated to \unsupported~ and vice versa). To illustrate with an example, $S_{\mathrm{original}}=$\{``claim'': ``5g helps covid-19 spread.'', ``evidence'': \textit{$evidence_{1}$}, ``label'': \unsupported \} is negated into $S_{\mathrm{negated}}=$\{``claim'': ``5g does \textbf{not} help covid-19 spread.'', ``evidence'': \textit{$evidence_{1}$}, ``label'': \supported \}.

\begin{table}[t]
\centering
\resizebox{\linewidth}{!}{%
\begin{tabular}{ccccc}
\toprule
Shot \# & Test Set & Models & Acc & F1-Macro \\ \midrule
\multirow{4}{*}{2} & \multirow{2}{*}{Original} & \roberta~ & 52.66\% & 34.29\%  \\
 &  & \pplgpt~ & \textbf{66.75\%} & \textbf{64.39\%}  \\ \cmidrule(lr){2-5} 
 & \multirow{2}{*}{\begin{tabular}[c]{@{}c@{}}Negation-\\ Augmented\end{tabular}} & \roberta~ & 46.75\% & 31.66\%  \\
 &  & \pplgpt~ & \textbf{ 52.98\%} & \textbf{50.99\%} \\ \bottomrule
\end{tabular}
}
\caption{Negation Analysis - Comparison between fine-tuned RoBERTa baseline classifier (\roberta~) and our perplexity-based classifier (\pplgpt~) on original \covidScientific~ dataset and its negation-augmented version in 2-shot setting.}
\label{table:negation_experiment}
\end{table}

\paragraph{Q1: Can the LM distinguish negation?} We use the new augmented Covid-Scientific dataset to investigate whether the LM manages to differentiate between the original-sentence $S_{\mathrm{original}}$ and negated-sentence $S_{\mathrm{negated}}$. The average of the absolute difference between the perplexities assigned to $S_{\mathrm{original}}$ and $S_{\mathrm{negated}}$ is $122$ and the maximum absolute difference value is $2800$.


\paragraph{Q2: Performance on negation-augmented dataset?} 
We evaluate the performance of the perplexity-based classifier (\pplgpt) on the ``negation-augmented" \covidScientific~dataset in reference to its original. Unsurprisingly, \pplgpt~ does experience a drop in performance of 13.77\% and 13.40\% in accuracy and F1-Macro, respectively. However, it still outperforms the fine-tuned \roberta~ baseline, the best performing baseline in the 2-shot setting, as shown in Table~\ref{table:negation_experiment}.

\subsection{Comparison with existing FEVER System in Few-shot Setting}
For all three tasks, we compare our perplexity models against different fine-tune baselines in Section \ref{experimental_results}. 
Unlike two newly proposed COVID-19-related tasks, FEVER is a well-established task studied by many existing works. In order to understand how our perplexity-based method compares against the literature, we conduct an additional experiment with the publicly available system from the runner-up team of the FEVER workshop, HExaF~\cite{yoneda2018ucl}. 

We fine-tune HExaF's veracity classification modules in few-shot settings. In the 2-shot settting, HExaF shows accuracy of 49.99\% and F1-Macro score of 33.33\%. In the 50-shot settting, it shows accuracy of 53.53\% and F1-Macro score of 49.27\%. 
In general, machine learning models require sufficient amounts of training data, and this "sufficient amount" normally differs depending on the model being used. However, as demonstrated earlier in our main experimental results (Section~\ref{experimental_results}), $2\sim50$ samples are insufficient data to properly train one of the winning fact-checking systems.



\subsection{Potential Application: Ranking of Candidate Claims for Fact-Checking }
Here, we discuss another way of leveraging the evidence-conditioned perplexity score. It can be used for prioritizing false-claim candidates for human fact-checkers, instead of doing hard prediction on the veracity of the given claims. 
By ranking the claims-to-be-fact-checked in descending order of perplexity, we can increase the chance that the first $k$ claims checked by a human fact-checker are \unsupported~ false claims. 
This will be beneficial since fact-checkers can efficiently allocate their time and resources on fact-checking claims that are more likely to be false and harmful to society. 

In Figure~\ref{fig:three graphs}, we compare the precision at the top-k (P@k) between the perplexity-based ranking and random-score-based ranking. We can view P@k to measure how many \unsupported~ pieces are prioritized in the first \textit{k} of the ranked claims. Across all datasets, perplexity-based ranking (blue marks) exhibits higher precision scores over random-score-based ranking (orange markers). Moreover, for both \covidScientific~and \covidSocial, our P@k is over 80\% for all \textit{k} values.

\section{Future Research Directions}
In this work, we conduct the FEVER experiments in a binary set-up to keep all the experimental settings consistent across all three datasets. However, the original FEVER task has three classes - Support, Refute, and Not Enough Info (NEI). Since the distinction between NEI and Refute cases is also an important problem, it would be important future work to extend our binary-class setting to the three-class setting. 

Moreover, we believe our method can easily be augmented into other existing approaches, for instance, leveraging the perplexity score in the final step of the FEVER fact-checkers as additional input. It would be a useful future direction to explore and discover the most effective way of incorporating the perplexity-based approach into other existing fact-checking systems.

\section{Conclusion}
\label{section:conclusion}
In this paper, we propose a novel way of leveraging the perplexity score from LMs for the few-shot fact-checking task. Through experimental analysis from an ablation study to the discussion of potential applications, we further explore and evaluate the capability of the perplexity score to act as an indicator of unsupported claims. We hope our proposed approach encourages future research to continue developing LM-based methodologies as well as the few-shot approach for fact-checking. By doing so, our community can move towards a data-efficient approach that is not constrained by the requirement of a large labeled dataset.

\bibliography{anthology,custom}
\bibliographystyle{acl_natbib}

\appendix

\end{document}